\documentclass[acmtog,nonacm]{acmart}
\setcopyright{none}
\makeatletter
\gdef\@authorsaddresses{}

\renewcommand{\@fnsymbol}[1]{%
  \ifcase#1\or \textdagger\else *\fi}
\makeatother

\AtBeginDocument{%
  }

\usepackage{subcaption}
\usepackage{tikz}
\usepackage{graphicx}
\usepackage{amsmath} 
\usetikzlibrary{patterns} 
\usepackage{multirow}
\usepackage{enumitem}
\usepackage{xcolor}         
\usepackage{ulem}           
\usepackage{booktabs}

\begin{document}

\title{
ShareVerse: Collaborative Video Generation for Shared World Modeling}

\author{JIAYI ZHU}
\affiliation{%
  \institution{Shanghai Jiao Tong University}
  \country{China}}
\email{larst@affiliation.org}

\author{JIANING ZHANG}
\affiliation{%
  \institution{Fudan University}
  \country{China}}

\author{YIYING YANG}
\affiliation{%
 \institution{Fudan University}
 \country{China}}

\author{WEI CHENG}
\affiliation{%
  \institution{StepFun}
  \country{China}}

\author{XIAOYUN YUAN}
\authornote{Corresponding Author.}
\affiliation{%
  \institution{Shanghai Jiao Tong University}
  \country{China}}
\email{cpalmer@prl.com}

\renewcommand{\shortauthors}{Trovato et al.}

\begin{abstract}
\noindent\hspace{13em}{\Large\textbf{Abstract}}\\[4pt]
The construction of large-scale, dynamic shared virtual worlds is a fundamental challenge in computer graphics and simulation. While recent large video generation models excel at synthesizing high-quality single-agent perspectives, they fundamentally lack the capability to collaboratively render a unified, spatiotemporally consistent environment across multiple agents.
In this paper, we present ShareVerse, a collaborative video generation framework that inherently supports consistent shared world modeling. Instead of relying on centralized global rendering, ShareVerse empowers distributed video generative agents to locally synthesize their surroundings while maintaining spatiotemporal consistency in overlapping fields of view.
Our framework introduces two core methodological advancements: (1) A cross-agent collaborative attention mechanism seamlessly integrated into the generation prior, enabling implicit routing of dynamic spatiotemporal features to resolve visual conflicts in overlapping regions. (2) A spatiotemporal memory retrieval system that anchors inter-agent long-range autoregressive generation, preserving environmental coherence and object identity over extended trajectories.
Extensive experiments demonstrate that ShareVerse dramatically outperforms existing  baselines in producing temporally stable, multi-view consistent environments, paving a new way for scalable virtual world synthesis.

\end{abstract}

\begin{teaserfigure}
  \includegraphics[width=\textwidth]{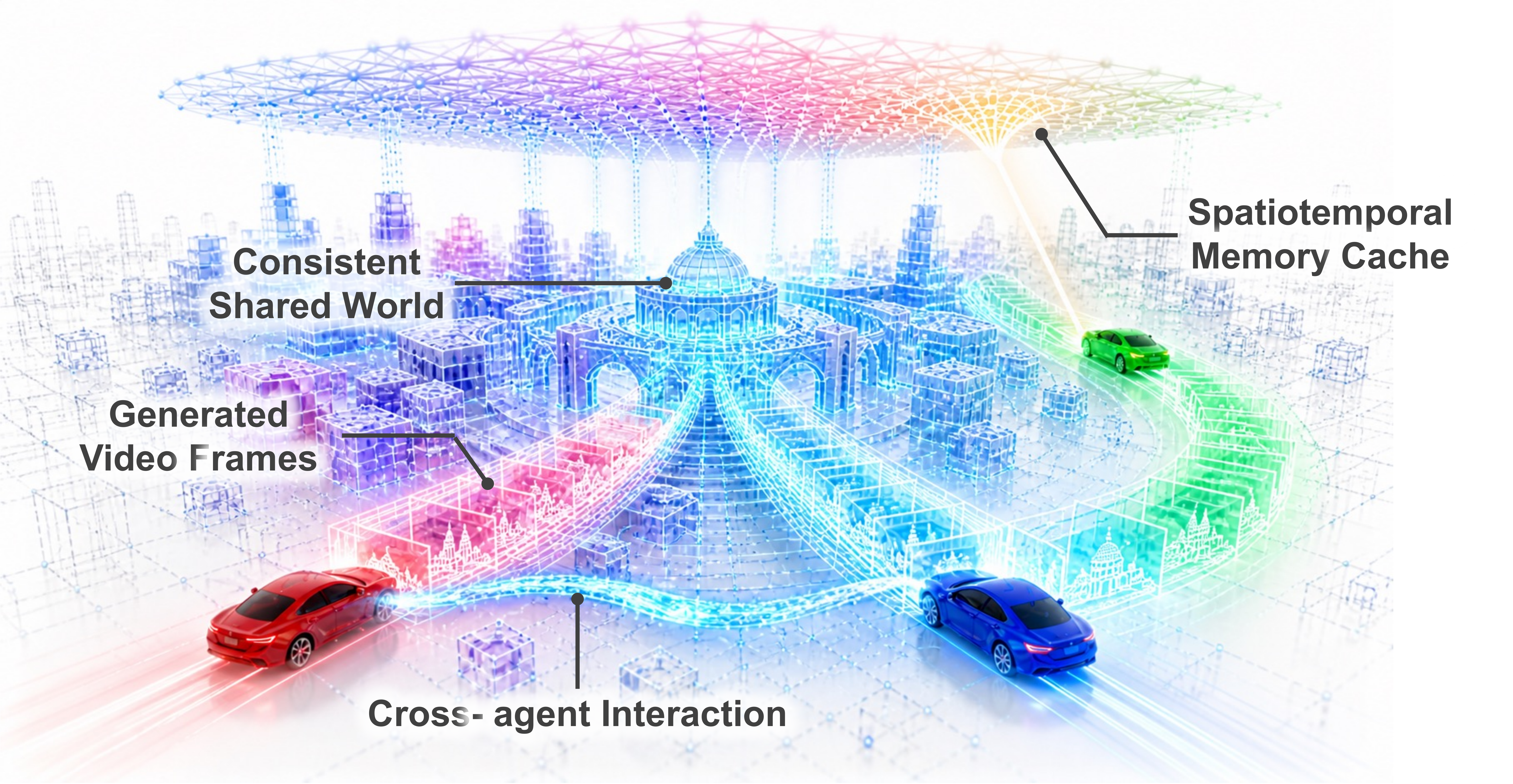}
  \vspace{-4mm}
  \caption{\textbf{ShareVerse: Collaborative Video Generation for Shared World Modeling.} ShareVerse empowers distributed agents to collaboratively synthesize a globally consistent virtual environment. We bridge isolated generative priors through two core mechanisms: (1) implicit {cross-agent interaction}, which resolves visual conflicts during concurrent exploration (red/blue vehicles); and (2) a global {Spatiotemporal Memory Cache}, which guarantees long-term environmental permanence during asynchronous revisitation (green vehicle).}
\label{fig:teaser}
\end{teaserfigure}

\maketitle

\section{Introduction}
Video generation models are rapidly emerging as a transformative paradigm for synthesizing virtual environments \cite{wan2025wan,zheng2024open,yang2024cogvideox,kong2024hunyuanvideo}. Compared to traditional computer graphics pipelines that rely on labor-intensive 3D asset authoring, these data-driven models show great potential to implicitly learn physical dynamics and structural priors directly from massive data. This capability paves a promising path toward visual "World Models" that simulate the physical world. However, a comprehensive world model cannot be limited to a single observer. To function effectively in complex applications, such as multiplayer games, autonomous driving, and multi-robot collaboration, the generated world must inherently accommodate multiple agents interacting within a shared space. Bridging the gap between isolated single-view video generation and multi-agent shared simulation has thus become a critical necessity.

In conventional graphics architectures, spanning from classic polygon meshes to recent neural rendering advancements like Neural Radiance Fields (NeRF) \cite{mildenhall2021nerf} and 3D Gaussian Splatting (3DGS) \cite{kerbl20233d}, rendering a "shared" virtual world is inherently guaranteed. These methods fundamentally rely on a unified 3D representation to model the entire environment before projecting to 2D via ray tracing or rasterization, ensuring that multiple agents naturally observe a strictly consistent scene. In stark contrast, modern video generation models operate as implicit synthesizers devoid of any centralized 3D anchor. Consequently, when multiple independent agents attempt to collaboratively simulate a dynamic scene from distributed perspectives, they hallucinate disparate realities in their overlapping fields of view. This visual divergence fundamentally shatters the spatiotemporal consistency required to sustain a cross-agent consistent shared virtual world. Therefore, the critical frontier lies in evolving these isolated generative priors into collaborative shared world models, harnessing their boundless generative capabilities while intrinsically maintaining the rigorous global consistency of a unified physical reality.

While recent advancements in generative world models \cite{zhu2025aether,yu2025context,he2025matrix,li2025hunyuan,wu2025video,gao2025longvie,xiao2025worldmem,li2025vmem,sun2025worldplay} have achieved remarkable temporal coherence for single-camera trajectories, they remain structurally ill-equipped for collaborative scenarios. Concurrently, emerging research in multi-view video synthesis \cite{xie2024sv4d,bai2024syncammaster,wu2025ic} fundamentally operates under a synchronous and single-agent paradigm, where multiple cameras merely capture a localized scene at the exact same moment. This approach drastically differs from the structural demands of a shared world. In a true shared environment, multiple independent agents often navigate the same spatial coordinates asynchronously. Constructing such a world requires robust long-range spatiotemporal memory to ensure that a location synthesized by one agent remains visually permanent and consistent when revisited by another agent in the future. Consequently, the ambitious pursuit of multi-agent shared world modeling remains largely unexplored within the community. It remains a significant challenge for current generative frameworks to seamlessly integrate these asynchronous trajectories to collaboratively render and sustain a globally consistent shared reality. 

To tackle these fundamental challenges, we propose ShareVerse, a collaborative video generation framework designed to synthesize consistent shared worlds. Driven by the need to ensure the spatiotemporal consistency, we ground our approach in a highly representative application domain: multi-agent autonomous driving simulation. In this context, each vehicle operates as a video generation agent. While our framework inherently enforces multi-view geometric consistency within each agent's local perspective, the core challenge lies in transitioning from these distributively generated videos to a globally coherent virtual environment.

To achieve global consistency without relying on explicit 3D geometry, ShareVerse introduces a spatiotemporal memory architecture driven by two integrated mechanisms. First, we establish an implicit synchronization pathway by embedding cross-agent collaborative attention directly into the generative rendering pipeline. This acts as a distributed consistency protocol, allowing independent agents to dynamically route spatial features and resolve visual conflicts in overlapping regions. Second, to define exactly what is exchanged in the absence of explicit 3D representations, we conceptualize the shared environment as a persistent cache of spatiotemporal neural tokens. We propose a memory retrieval system where agents dynamically query and fetch historical tokenized scenes from a collaborative pool based on relative spatial proximity. Instead of relying on explicit 3D representations, agents directly share these retrieved memory tokens via cross-agent attention to condition their current synthesis. By anchoring the generation process to these shared historical features, ShareVerse effectively prevents scene drifting. This ensures that the virtual environment remains visually and structurally permanent, even when different agents navigate the same location at different times. 
Extensive experiments demonstrate that ShareVerse dramatically outperforms existing independent generation baselines in producing temporally stable and cross-agent consistent shared environments.

\begin{figure}[t]
	\centering
\includegraphics[width=1\linewidth]{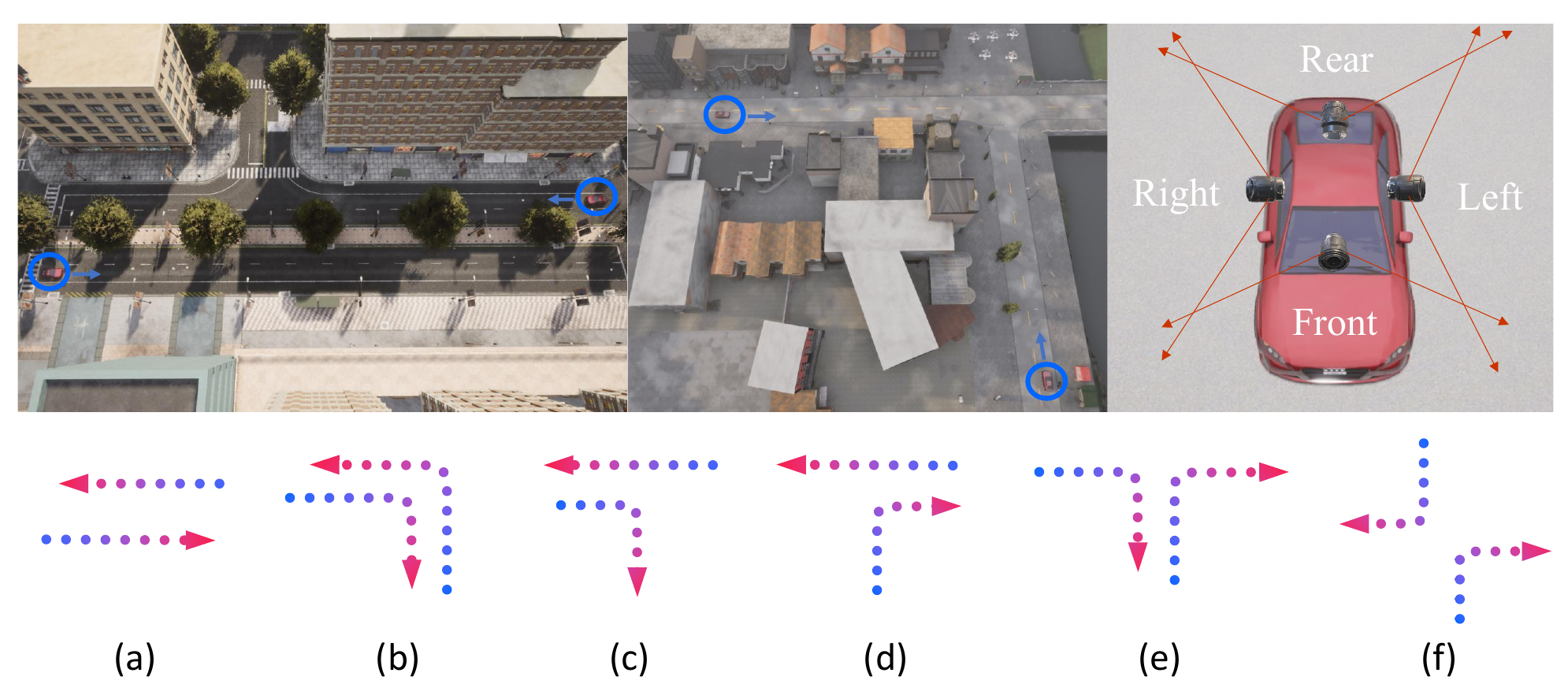}
    \vspace{-6mm}
     \caption{{Illustration of the dataset construction process}. We build the multi-agent synchronized training dataset by rendering in CARLA. We equip each agent with four cameras (front, rear, left, and right) and choose six main trajectory pairs, as shown in (a) -- (f). Each pair enables interaction between agents.}
     \label{fig:dataset}
     \vspace{-4mm}
\end{figure}

\vspace{-3mm}
\section{Related Work}
\subsection{Video Generation Models}
Diffusion models \cite{ho2020denoising,lipman2022flow,song2020score} have emerged as the dominant state-of-the-art method in video generative modeling. Researchers have adapted diffusion model architectures to the temporal domain, enabling the generation of high-quality video clips containing tens to hundreds of frames \cite{wan2025wan,zheng2024open,yang2024cogvideox,kong2024hunyuanvideo,chen2025skyreels}. Furthermore, through techniques such as ControlNet\cite{zhang2023adding} and Distillation \cite{frans2024one,geng2025mean}, related methods have achieved further advancements in generation controllability and efficiency.

Traditional video models are limited to producing videos of a fixed frame length.
Recent autoregressive approaches \cite{wu2025video, yu2025context, xiao2025worldmem} include generating new frames given a set of historical generated frames as context.
Others \cite{chen2024diffusion, huang2025self} modify the diffusion objective by assigning independent noise levels for each frame during training, allowing AR inference. These methods can be used for infinite-length generation by utilizing a sliding window context.

\subsection{Multi-View Models}
Current research in multi-view video generation primarily focuses on object-level or small-range 4D synthesis. SV4D \cite{xie2024sv4d} generates multiple novel view videos by leveraging both 3D priors in multi-view image generation models and motion priors in video generation models. SynCamMaster \cite{bai2024syncammaster} designs a
multi-view synchronization module to maintain appearance and geometry consistency across different viewpoints. IC-world \cite{wu2025ic} enables parallel and coherent multi-view generation through pixel-wise coupling and instruction-based in-context prompting.
These are fundamentally distinct from the work we present, which focuses on multi-agent long-range world modeling, with the goal of realizing the dynamic interaction and collaborative generation of multiple agents to construct a shared world.

\subsection{World Models}
Driven by discrete or continuous action signals, world models such as \cite{he2025matrix,che2024gamegen,li2025vmem,xiao2025worldmem,bar2025navigation,li2025hunyuan,mao2025yume,sun2025worldplay} aim to navigate and interact with virtual environments. 
To ensure the geometric consistency of generated videos, existing works can be categorized into two types: explicit reconstructions and implicit conditioning. \cite{wu2025video,li2025vmem,cao2025uni3c,ren2025gen3c,yu2025trajectorycrafter} explicitly reconstruct 3D representations and render frames based on these representations as conditions for new video generation. In contrast, \cite{xiao2025worldmem,yu2025context,sun2025worldplay}  directly retrieve relevant historical frames as conditions through field-of-view (FOV) overlap, demonstrating efficiency and scalability.
However, traditional world models struggle to address shared world modeling generated by multiple agents.


\begin{figure*}[t]
	\centering
\includegraphics[width=1\textwidth]{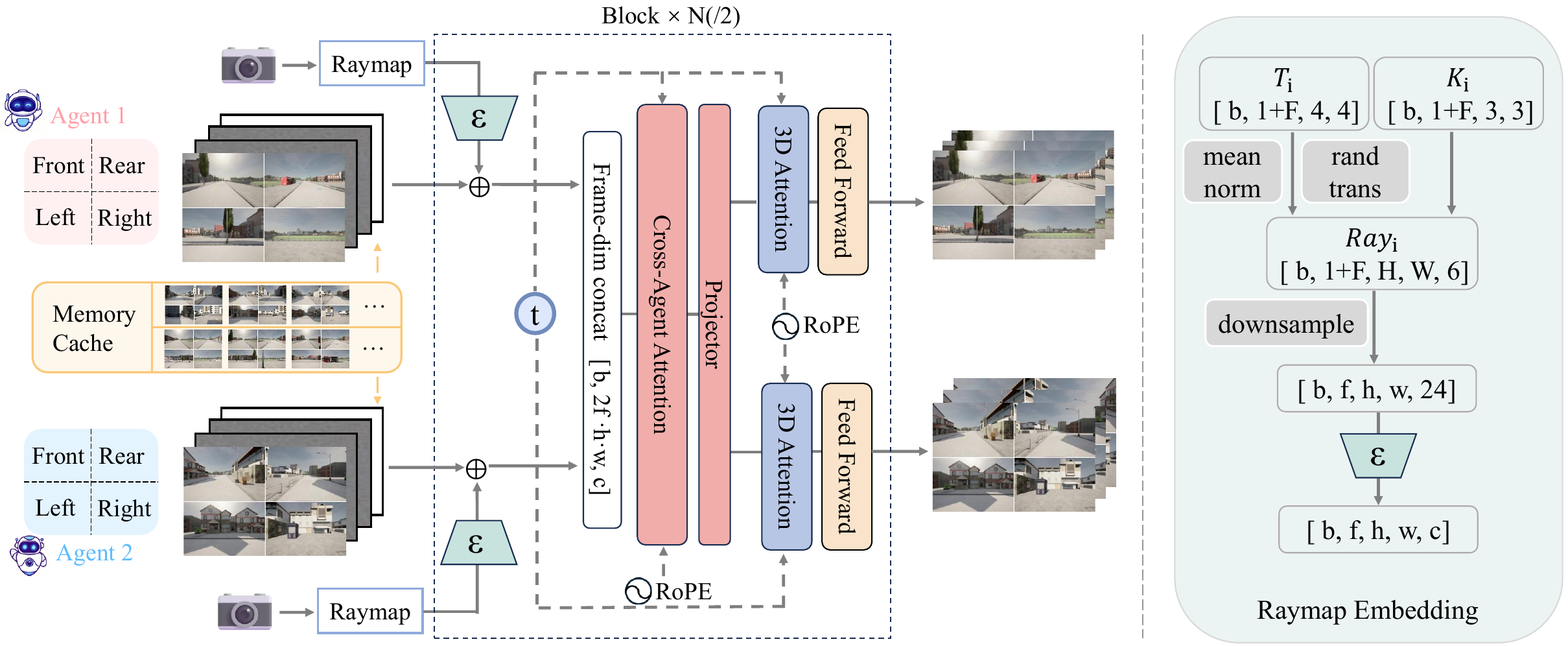}
    \vspace{-6mm}
     \caption{{Method overview}. Given one image from each of two agents, where four views are concatenated to depict a scene, \textbf{ShareVerse} performs a prediction task to generate future videos conditioned on the camera trajectories from users. In the generation process, two agents explore the world, exchange captured visual information, and perceive each other’s positions.}
     \label{fig:method}
     \vspace{-4mm}
\end{figure*}

\section{ShareVerse Methodology}
\subsection{Problem Formulation and Framework Overview}

Synthesizing shared virtual environments for multiple agents constitutes a formidable challenge in generative simulation. To systematically address this within a rigorous and evaluable domain, we instantiate the ShareVerse framework within the context of multi-agent autonomous driving. In this paradigm, each vehicle functions as a distributed generative node. Mimicking the omnidirectional perception systems of modern intelligent vehicles, each agent employs an orthogonal four-view camera configuration (front, rear, left, and right). This sensory arrangement provides the continuous panoramic coverage strictly required for high-fidelity local scene rendering.

The core generative engine of ShareVerse leverages the architectural foundation of CogVideoX \cite{yang2024cogvideox}. Specifically, a 3D Variational Autoencoder (VAE) \cite{kingma2013auto} compresses raw multi-view video streams temporally and spatially into a compact latent space.
A Diffusion Transformer (DiT) \cite{peebles2023scalable} then iteratively denoises these latents to synthesize video frames. While this baseline model provides a powerful architecture for high-quality video synthesis, it intrinsically operates as a single-perspective generator. It lacks both the structural geometric awareness and the cross-agent synchronization mechanisms necessary to sustain a globally coherent shared reality.

To enable collaborative video generation for a shared world model, ShareVerse integrates four specialized components:
\begin{itemize}
    \item Intra-agent spatial awareness: We establish robust local perception by spatially concatenating the four orthogonal views and injecting explicit geometric raymaps, thereby granting each independent agent strict spatial awareness (Sec. 3.2).
    \item Cross-agent interaction: We introduce a cross-agent collaborative attention protocol that routes implicit spatial features to enforce inter-agent visual consistency, resolving the inevitable visual conflicts that arise when multiple agents render overlapping regions (Sec. 3.3).
    \item Spatiotemporal memory retrieval: We design a spatiotemporal memory retrieval system that bypasses explicit 3D representations to guarantee long-term environmental permanence, preserving this consistency across asynchronous temporal trajectories (Sec. 3.4).
    \item Simulation-based dataset: We detail our procedural data synthesis pipeline to satisfy the massive scale and rigorous multi-view synchronization required to train and optimize this collaborative framework (Sec. 3.5).
\end{itemize}

\subsection{Intra-Agent Spatial Awareness}
In multi-agent collaborative video generation, a fundamental prerequisite is that each independent agent must possess explicit awareness of its position and pose within the global environment. An agent cannot effectively coexist or interact within a shared world without first establishing its spatial localization. To endow the baseline video generation model with this crucial spatial awareness, we employ two targeted approaches: a tiled surround-view input formulation and a dense ray-based camera conditioning mechanism.

First, to achieve comprehensive perception of the immediate surroundings, we emulate the 360-degree surround-view monitoring systems equipped on modern intelligent vehicles \cite{hecker2018end,petrovai2022semantic,kumar2023surround,lin2025one}. This is realized through a four-camera configuration covering the front, rear, left, and right directions (Fig.~\ref{fig:dataset}). Rather than processing these streams independently, we spatially concatenate the four views into a single $2 \times 2$ tiled frame for training. This unified input strategy requires no modifications to the underlying video model. It significantly reduces memory overhead while inherently forcing the network to aggregate a comprehensive and structurally coherent environmental context.

Second, we mathematically anchor these latent features to the global environment by explicitly conditioning the generative process on camera trajectories. Formally, given the intrinsic matrix $K$ and camera pose $T$ (comprising rotation $R$ and translation $t$), we translate the physical camera parameters into dense geometric raymaps. To mitigate extreme coordinate values that could destabilize training, the translation vector $t$ is first mean-normalized. Specifically, for each of the four concatenated views, we utilize $K$ to compute the local camera ray direction, which is then transformed by $R$ into the world-coordinate ray direction $r_d$. The ray origin $r_o$ is set to the normalized translation vector. These parameters are concatenated ($[r_d, r_o]$) to formulate a base 6-channel geometric raymap. To rigorously align with the latent representation without redundant computational overhead, these raymaps are computed directly at the reduced spatial resolution ($h \times w$) of the VAE latents. Furthermore, because the 3D VAE temporally compresses every four consecutive video frames into a single latent frame, the temporal length of the raymaps must be identically reduced to match the latent dimension $f$. To achieve this synchronization without losing geometric information, we pack the 6-channel raymaps of the four corresponding original frames along the channel dimension. This operation yields a spatiotemporal geometric tensor $\mathit{Ray} \in \mathbb{R}^{b \times f \times h \times w \times 24}$. A dedicated raymap encoder then projects this tensor to match the hidden dimensions of the DiT. Ultimately, this encoded geometric condition is added element-wise to the video latent features every two DiT blocks, explicitly anchoring the generation process to precise 3D spatial coordinates.

While these spatial conditioning strategies enable geometrically grounded local video synthesis, individual agents remain completely isolated. To construct a globally shared world, we subsequently detail our dedicated mechanisms for inter-agent visual consistency and long-term environmental persistence.

\subsection{Cross-Agent Interaction}
To synthesize a globally coherent shared environment and prevent structural hallucinations in overlapping regions, we introduce a Cross-Agent Collaborative Attention module. For clarity, we use a two-agent scenario as an example to illustrate this synchronization process. The effectiveness of this module relies fundamentally on the geometric anchoring established in the previous section. Because the latent features $F_i$ (where $i \in \{1, 2\}$) are explicitly conditioned on geometric raymaps, they inherently operate within a shared 3D coordinate system. Consequently, applying attention across these agent features forces the network to align and synchronize representations that occupy the exact same physical space, effectively resolving spatial conflicts.The module is designed as a streamlined pipeline of concatenation, attention, and splitting, ensuring that input and output dimensions remain strictly identical for seamless integration into the iterative DiT blocks. First, the input features $F_1$ and $F_2$ are concatenated along the temporal dimension to form an aggregated feature sequence. Because this concatenation operation doubles the temporal length, we apply Rotary Position Embedding (RoPE) \cite{su2024roformer} to properly index the frames across the newly aggregated sequence. This sequence is then processed by the attention mechanism to exchange information and enforce consistency. Finally, the output is projected via a zero-initialized linear layer, split back into individual agent features, and applied as a residual update. These globally synchronized features are subsequently fed into the standard DiT blocks.This cross-agent collaborative process, coupled with the initial geometric raymap injection ($\mathcal{E}_c(\mathit{Ray}_i)$), is formally defined as follows:


\begin{equation}
F_i^\text{in} = F_i + \mathcal{E}_c(\mathit{Ray}_i), F_\text{agg} = \text{RoPE}\!\left( \left[F_1^\text{in}, F_2^\text{in}\right] \right),
\end{equation}
\begin{equation}
F_i^\text{sync} = F_i^\text{in} + \text{Split}\!\left(\text{Proj}\!\left(\text{Attn}_\text{cross}(F_\text{agg})\right)\right)_{\!i}, F_i^\text{out} = \text{DiT}(F_i^\text{sync}).
\end{equation}

\subsection{Spatiotemporal Memory}
While cross-agent attention resolves visual conflicts during concurrent generation, a shared world intrinsically demands long-term structural consistency. When an agent revisits a previously explored location, the synthesized scene must remain stable to prevent catastrophic forgetting. To achieve this global permanence without relying on explicitly maintained 3D representations, we propose a spatially-driven memory retrieval mechanism for autoregressive video generation.

During navigation, each agent continuously archives its synthesized frames and corresponding physical coordinates into a spatial memory cache. To generate the subsequent video chunk, the agent does not merely fetch the temporally closest frames. Instead, it samples key spatial coordinates along its future trajectory and retrieves historical frames from the cache that are physically proximate to these upcoming locations. Specifically, given 4 uniformly sampled positions on the future trajectory of agent $a$: $\mathbf{p}_a^{(1)}, \mathbf{p}_a^{(2)}, \mathbf{p}_a^{(3)}, \mathbf{p}_a^{(4)}$ and memory cache of agent $b$: $\mathcal{M}_b = \{ (\mathbf{f}_b^i, \mathbf{p}_b^i) \}_{i=1}^{N}$, the retrieval mechanism can be formulated as:
\begin{equation}
\mathcal{M}_{a \leftarrow b} = \{ ( \mathbf{f}_b^{i_j}, \mathbf{p}_b^{i_j} ) \,\big|\, i_j = \arg\min_{i} \| \mathbf{p}_a^{(j)} - \mathbf{p}_b^i \|_2,\ j=1,2,3,4 \},
\end{equation}
where $\mathcal{M}_{a \leftarrow b}$ denotes the historical frame-position pairs retrieved by agent $a$, and $\mathbf{f}_b$ and $\mathbf{p}_b$ represent the historical frames and trajectory positions of agent $b$, respectively. Conditioning the generative process on these spatially adjacent memories natively forces the new visuals to align perfectly with established environmental structures. Furthermore, we apply a unidirectional attention mask within the transformer blocks. While the newly generated frames actively query the memory frames for spatial guidance, the memory frames are restricted to self-attention. This architectural design explicitly prevents the clean historical features from being corrupted by the highly noised latents during the iterative denoising process.

To train this retrieval-augmented architecture, we design a chunk-wise conditioning protocol. Specifically, each training chunk is explicitly configured to 33 frames, comprising 29 frames to be generated and 4 retrieved memory frames. To bridge the domain gap between the clean ground-truth frames used during optimization and the imperfect historical frames encountered during actual inference, we inject controlled noise perturbations into the initial conditioning frame. At inference time, the model queries the spatial memory cache based on its planned trajectory and autoregressively synthesizes continuous video chunks, ensuring the shared virtual world remains geometrically anchored and temporally permanent throughout the simulation.

\begin{figure}[t]
	\centering
\includegraphics[width=1\linewidth]{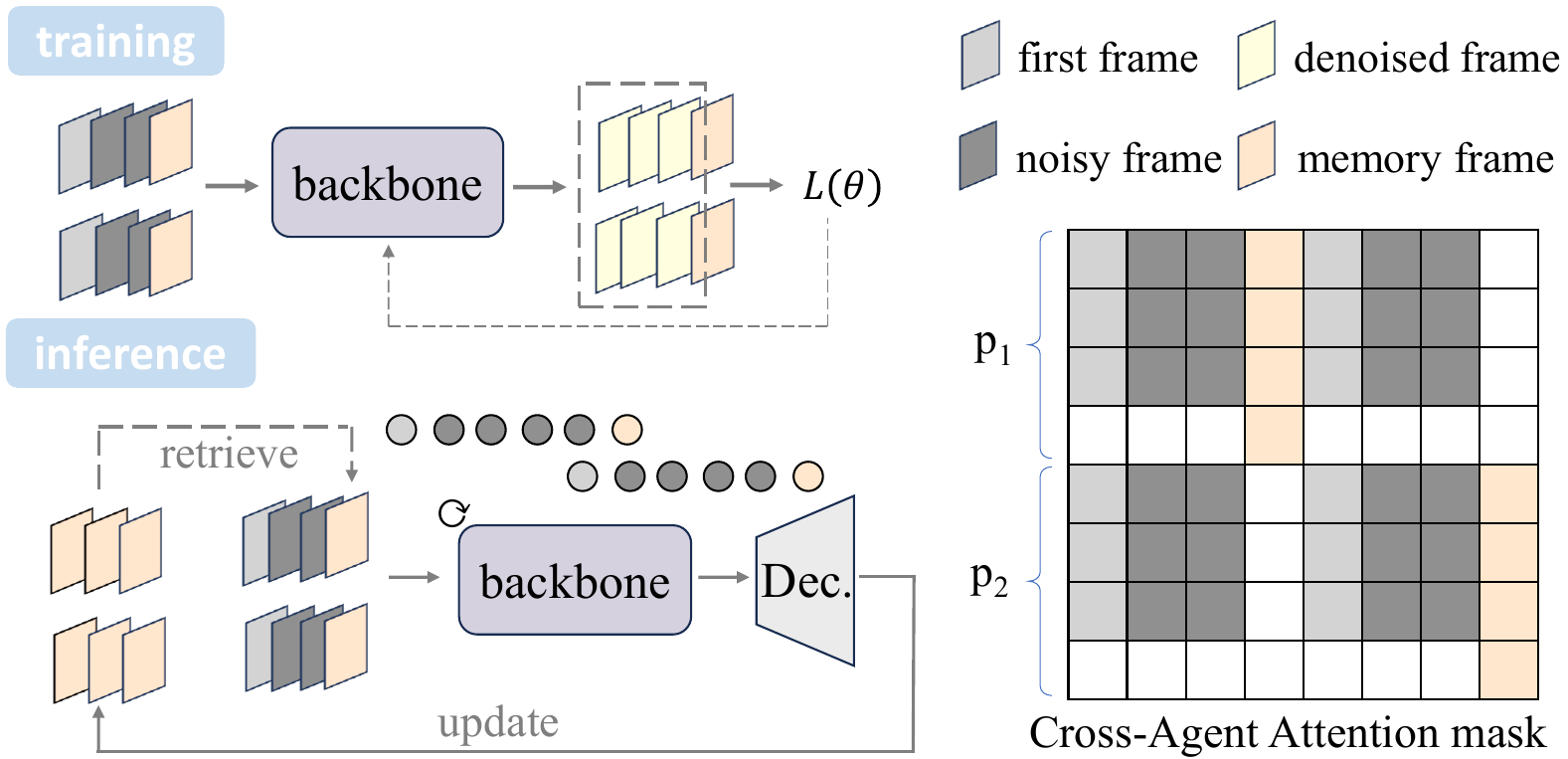}
    \vspace{-6mm}
     \caption{{Pipeline overview}. By retrieving memory frames from memory cache and concentrating them with frames to be generated to guide generation, our approach ensures long-term consistency in shared world simulation.}
     \label{fig:pipeline}
 \vspace{-4mm}
\end{figure}

\subsection{Multi-Agent Dataset Synthesis}
Training the ShareVerse framework requires perfectly synchronized video sequences captured by multiple distributed agents within the exact same spatiotemporal context. To satisfy this strict requirement, we develop a procedural data generation pipeline utilizing the CARLA simulation engine \cite{dosovitskiy2017carla}.
Our dataset construction follows a structured protocol designed to systematically maximize both scene and interaction diversity. First, to ensure scene diversity, data collection is conducted across various urban topologies under distinct weather conditions, providing a comprehensive atmospheric background. Second, to enrich interaction diversity, we design specific interactive trajectory modes among multiple agents, such as head-on encounters and intersection crossings (Fig. \ref{fig:dataset}). We further enhance this behavioral richness by applying random spatial shifts to the start and end points of each basic trajectory. Finally, to simulate comprehensive local perception, each agent is equipped with four synchronized orthogonal cameras (front, rear, left, and right). This configuration efficiently captures multiple sets of multi-view RGB videos synchronously, covering an unbroken 360° local field of view for every distributed agent.
Ultimately, this streamlined pipeline yields a large-scale, long-horizon dataset containing over 15,000 perfectly synchronized multi-agent video sequences, with each continuous sequence lasting approximately 90 frames.

\begin{figure*}[t]
	\centering
\includegraphics[width=0.98\linewidth]{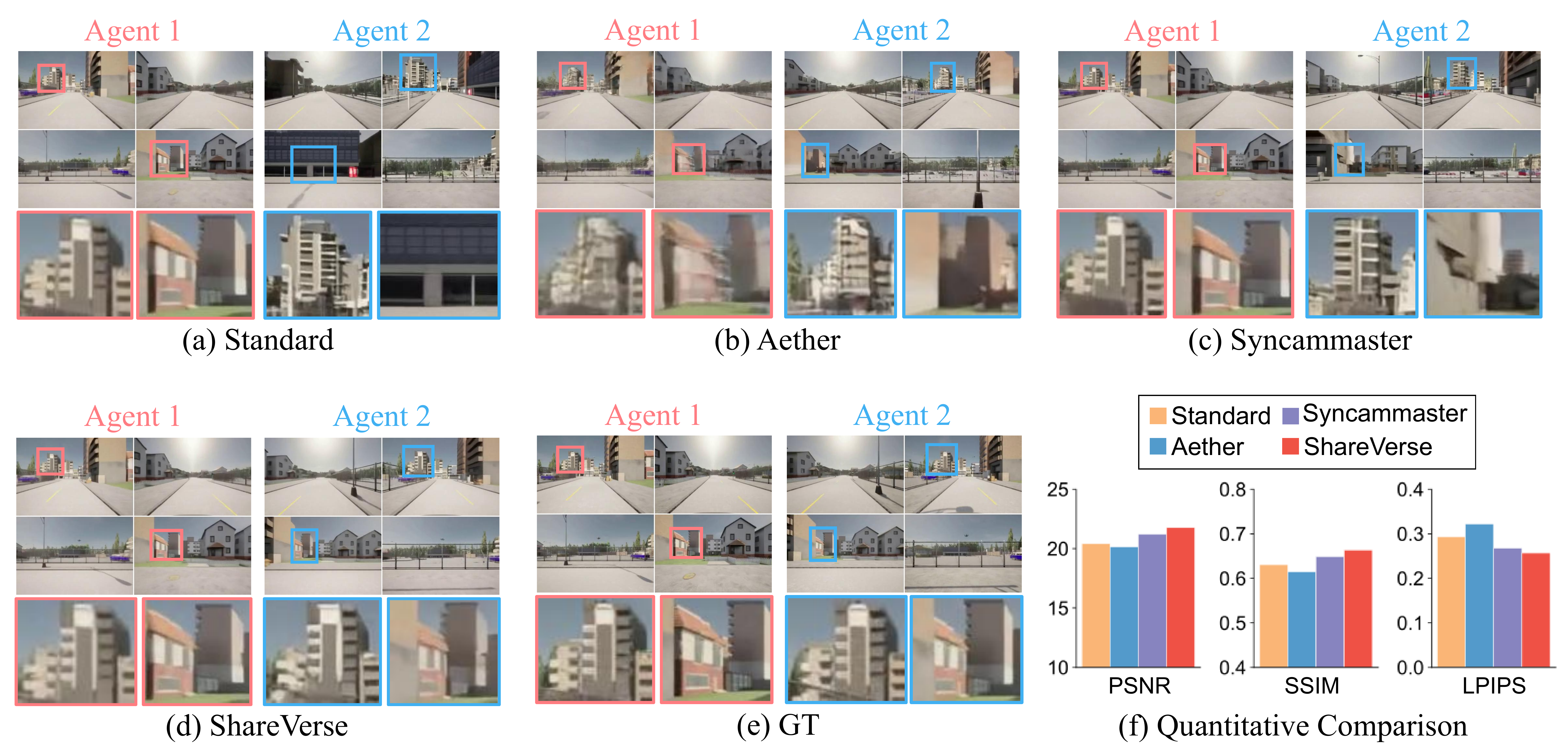}
    \vspace{-4mm}
\caption{\textbf{Evaluation of cross-agent visual consistency.} 
(a-e) Generated 4-view video frames and zoomed-in local crops are compared across baselines and our method. 
The red and blue boxes denote the identical architectural structure observed by Agent 1 (left) and Agent 2 (right), respectively. 
ShareVerse accurately preserves structural integrity across distinct viewports and closely aligns with the Ground Truth. 
(f) Quantitative metrics further confirm that ShareVerse consistently outperforms baselines across PSNR, SSIM, and LPIPS.}
     \label{fig:comparison}
     \vspace{0mm}
\end{figure*}

\section{Experiments}
\subsection{Implementation Details}
\label{sec:5.1}
We implement the ShareVerse framework based on the CogVideoX-5B-I2V architecture \cite{yang2024cogvideox}. The optimization process is conducted in two primary stages. First, we train a full-sequence base model for 15,000 steps, generating 49-frame videos at a resolution of $480 \times 720$. Subsequently, we adjust the temporal sequence length to 33 frames and further fine-tune the complete architecture for an additional 16,500 steps, explicitly incorporating the spatiotemporal memory mechanism and unidirectional attention masking. In both stages, we adopt the loss function of a standard latent diffusion model \cite{rombach2022high} :
\begin{equation}
\mathcal{L}(\theta) = \mathbb{E}_{t,x_0,\epsilon} \|\epsilon_\theta(x_t, t) - \epsilon\|_2^2.
\end{equation}
The entire training pipeline is executed on 8 NVIDIA A100 GPUs with a learning rate of $2 \times 10^{-5}$ and a global batch size of 24. To preserve pre-trained priors, the cross-agent attention blocks are initialized with the weights of the standard CogVideoX DiT blocks, while the raymap encoders and linear projectors are zero-initialized. 

\subsection{Baselines and Evaluation Metrics} 
To verify the capability of ShareVerse in synthesizing a shared world, we construct a validation dataset comprising unseen complex urban scenes. A key challenge in this evaluation is that independent agents initiate their trajectories with zero field-of-view (FOV) overlap. Overlapping regions only emerge dynamically during the generation process, often occurring asynchronously when one agent visits a location previously explored by another at a different time interval. Thus, the evaluation focuses on the model's ability to maintain long-term spatiotemporal consistency under these sparse and delayed interaction constraints.

\subsubsection{Baselines}
To elucidate the fundamental differences in multi-agent processing, we formalize the latent video representation of a single agent as a tensor $\mathbf{Z} \in \mathbb{R}^{f \times h \times w \times c}$, where $f, h, w$, and $c$ denote the temporal frame, height, width and channel dimensions, respectively. We compare ShareVerse against three representative paradigms based on how they manipulate these dimensions for cross-agent communication. For a fair comparison, all baselines have been trained on our dataset to ensure proper alignment with our specific autonomous driving scenarios.
\begin{itemize}
    \item Standard: utilizing the standard CogVideoX-I2V architecture without any tensor integration. This represents the default performance where agents navigate the environment in complete isolation, serving as a lower bound.
    \item Aether \cite{zhu2025aether}: concatenating the multi-agent tensors along the channel dimension $c$. The attention mechanism processes this composite tensor as a single entity, forcing the network to implicitly entangle distinct viewports within a shared latent space. 
    \item SyncamMaster \cite{bai2024syncammaster}: concatenating multi-agent tensors along the frame dimension $f$ at the same timestep. The attention module subsequently operates across this rearranged temporal sequence. While this mechanism explicitly aligns views at identical timesteps, it enforces a rigid temporal lock.
\end{itemize}
    
\subsubsection{Evaluation Metrics.} 
To comprehensively evaluate the ShareVerse framework, we design an evaluation protocol covering two complementary dimensions: reference-based geometric reconstruction, cross-agent spatiotemporal consistency.

First, for the reference-based evaluation, we compute traditional image reconstruction metrics including PSNR, SSIM \cite{wang2004image}, and LPIPS \cite{zhang2018unreasonable} against the ground truth (GT) videos in physically overlapping regions. 

Second, the core objective of structural alignment and cross-agent visual consistency is evaluated via a proposed reprojection protocol. This metric quantifies the geometric consistency of the shared world by measuring the alignment of synthesized physical structures across disparate agent trajectories, ensuring that multiple observers perceive a strictly shared environment.

To ensure a fair and deterministic comparison for diffusion models, we strictly fix the random seeds, text prompts, and initial conditioning frames across all inference steps.


\subsection{Rationality of Four Views for a Single Agent} 
\begin{figure*}[h] 
    \centering
    \includegraphics[width=\linewidth]{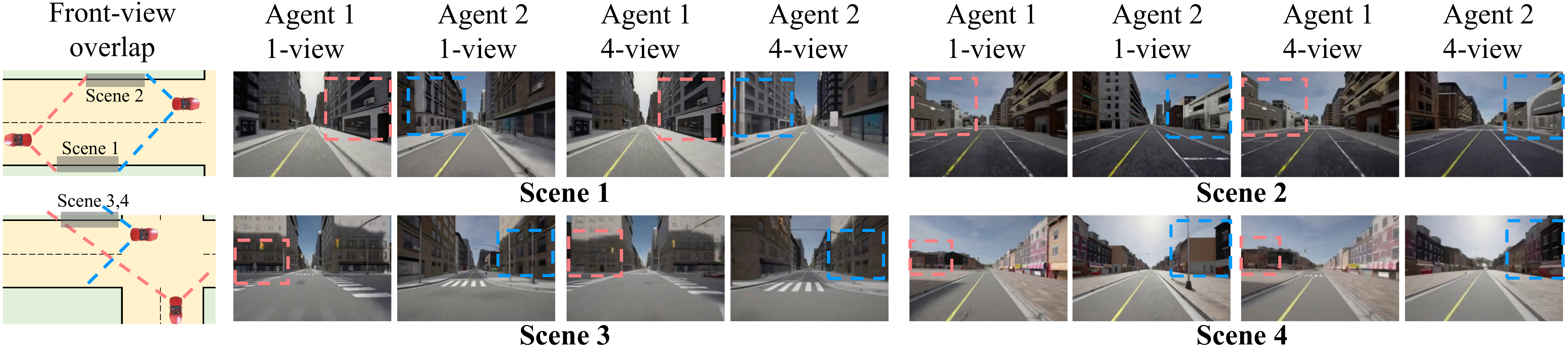}
    \vspace{-6mm} 
    \caption{\textbf{Front-view visualization for different training paradigms.} We compare the results of agents trained solely on front-view videos and agents trained on 4-view videos. The frames are sampled by the front view of two agents at different timesteps, but the overlap regions (gray shadow), which are enclosed by the red and blue dashed boxes on images, are supposed to be identical. Obviously, agents with 1 view generate visual hallucinations within overlap regions while agents with 4 view achieve consistency.
}
  \vspace{0mm}
\label{fig:single_view_comparison}
\end{figure*}

\begin{figure*}[h] 
    \centering
\includegraphics[width=\linewidth]{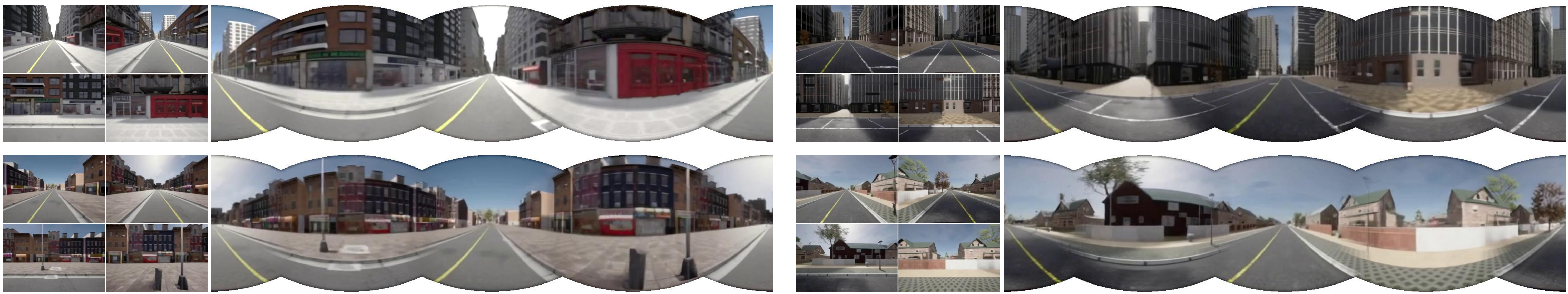}
    \vspace{-6mm}
  \caption{\textbf{Panoramic visualization for analyzing single-agent view coherence.} For each example, we display the generated 4-view video frames (left) and the corresponding 360-degree panorama stitched from these frames (right). By projecting the four distinct viewports onto a panoramic canvas using the exact camera parameters, it is evident that the architectural facades and distant horizons exhibit remarkable seamlessness across camera boundaries.}
    \vspace{0mm}
    \label{fig:pano}
\end{figure*}

Before examining multi-agent interactions, we verify the rationality of adopting four-view input for a single agent during generation process. As shown in Fig. \ref{fig:single_view_comparison}, training on front-view videos leads to severe generation instability and fails to maintain consistency within overlapping front-view regions, whereas panoramic coverage endows agents with a global field of view to perceive surroundings and relative positions, enabling superior spatial reasoning and global consistency.
We further stitch the generated frames into panoramas using the exact camera parameters provided as inputs to the video generation model (Fig. \ref{fig:pano}). Notably, while minor misalignments on the near-field ground naturally occur due to the perspective parallax introduced by the non-concentric multi-camera configuration, the architectural facades and distant horizons exhibit remarkable seamlessness. This consistent integration of vertical topologies across view boundaries verifies that ShareVerse enforces an accurate 360-degree spatial constraint, effectively proving the viability of our unified generation strategy in constructing a singular and structurally coherent local environment.

\subsection{Quantitative Analysis of Cross-Agent Consistency}
\label{cross-agent consistency}
To quantify cross-agent geometric consistency, we adopt a reprojection-based metric. We first match spatially adjacent view pairs from different agent trajectories. Specifically, for each scene, we select the pair of views from two different vehicles whose camera poses are the most similar, considering both spatial proximity and camera orientation consistency. The similarity is determined based on the estimated camera poses. For each selected pair, we independently estimate the depth maps and camera poses using Pi3 \cite{wang2025pi}. The evaluation follows a direct step-wise projection process. For a pixel in the source view, we use its estimated depth to project it into the target view’s image plane. At this projected location in the target view, we then use the corresponding target depth to project the point back to the original source view. The Euclidean distance between the initial pixel and the final reprojected location defines the reprojection error. This process is applied symmetrically and the results are averaged.

A point is considered valid if its reprojection error is below 10 pixels and it remains within the image boundaries (excluding sky regions). We report the {Mean Reprojection Error} over valid points and {Mean Completeness} (the ratio of valid points). As summarized in Table~\ref{tab:cross_exp}, ShareVerse achieves the lowest mean error of 1.6286 and the highest completeness of 0.6206. These results demonstrate that our framework effectively minimizes geometric discrepancies and maintains superior consistency in the synthesized shared world compared to existing baselines.

\begin{table}[htbp]
\centering
\begin{tabular}{lcc}
\hline
Methods & Mean Error & Mean Completeness \\
\hline
Standard & 3.3174 & 0.3935 \\
Aether & 2.6217 & 0.5404 \\
Syncammaster & 2.2783 & 0.5713 \\
Ours & \textbf{1.6286} & \textbf{0.6206} \\
\hline
\vspace{-4mm}
\end{tabular}
\caption{Quantitative analysis of cross-agent consistency}
\label{tab:cross_exp}
\end{table}
\vspace{-8mm}

\subsection{Qualitative Analysis of Cross-Agent Consistency} 
\begin{figure*}[h]
	\centering
\includegraphics[width=1\linewidth]{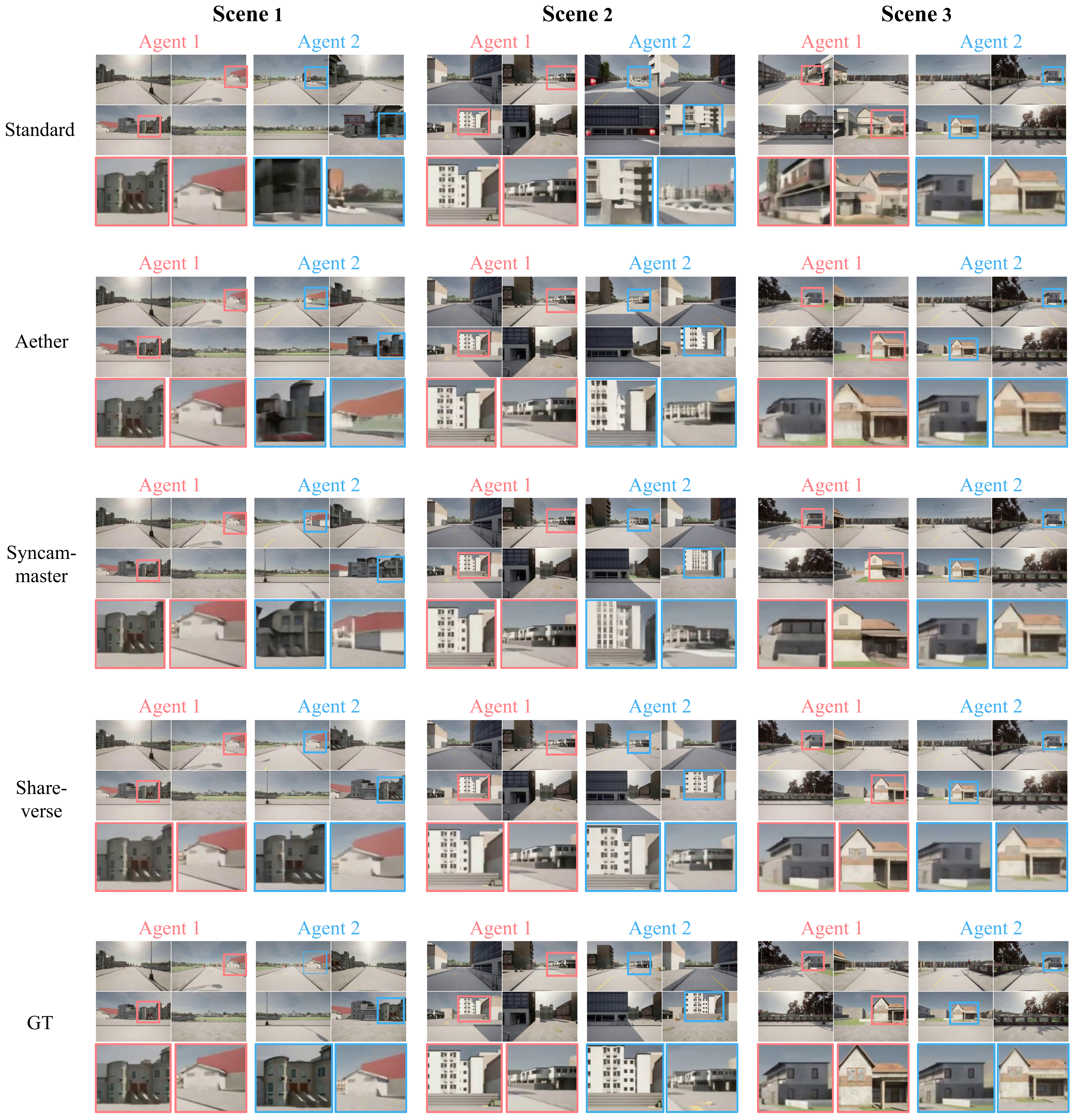}
    \vspace{-6mm}
    \caption{More visual results compared with baselines. }
 \label{Fig:more}
\end{figure*}

We first compare ShareVerse against multiple baseline methods in Fig. \ref{fig:comparison}. 
The local crops provide clear evidence of cross-agent visual consistency. 
It is evident that the baselines (Standard, Aether, and Syncammaster) produce severe artifacts and fail to synthesize the target building consistently across the two distinct agents. 
Conversely, ShareVerse accurately renders fine-grained architectural details, maintaining a stable shared environment. 
This high level of structural fidelity demonstrates that our method significantly outperforms existing approaches and achieves generation quality comparable to the GT. More results are in Fig. \ref{Fig:more}

\begin{figure*}[h] 
\centering
\includegraphics[width=\linewidth]{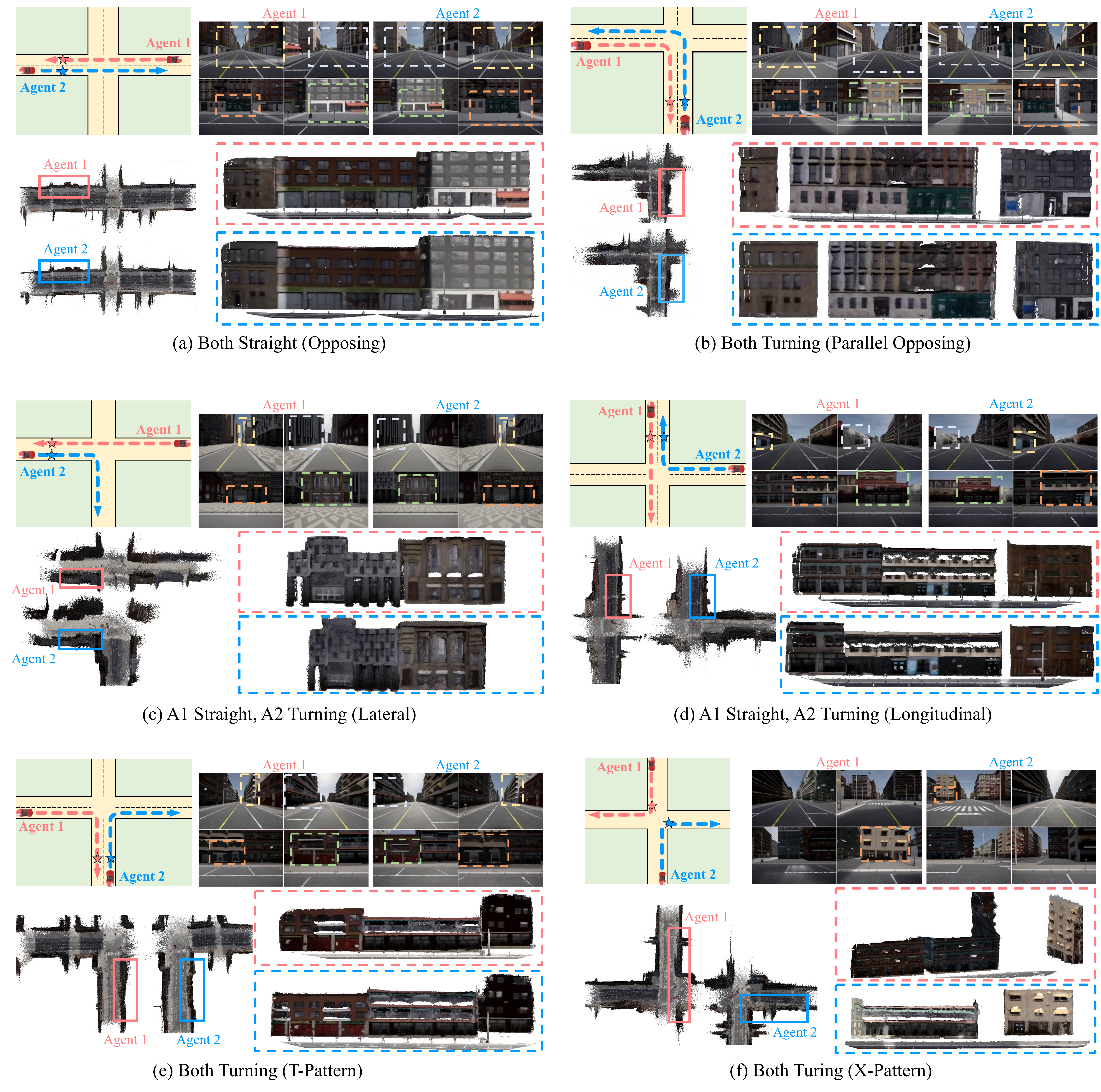}
\vspace{-8mm}
\caption{\textbf{Qualitative analysis of spatiotemporal and geometric consistency across diverse interactive trajectories.} 
We evaluate ShareVerse under six representative two-agent scenarios. 
For each scenario panel, we visualize the agent trajectories (top-left), the generated 4-view video frames (top-right), and the 3D point clouds reconstructed independently from each agent (bottom). 
Within the generated frame grids, color-coded dashed boxes denote identical physical regions observed from disparate viewing angles. 
For example, in the opposing straight paths scenario (a), the dashed boxes highlight specific storefront facades, illustrating that both agents render identical street views despite facing opposite directions. 
Similarly, red and blue dashed boxes in the 3D point clouds highlight the structural consistency of building geometries across viewports. 
These 3D reconstructions accurately reflect the architectural details found in the generated 2D video frames, while the point clouds from the two agents exhibit an accurate geometric alignment with each other. 
Notably, even in the highly challenging X-junction crossing (f), where spatial overlap is transient and complex, our model maintains an accurate shared world. 
This is directly evidenced by the 3D reconstruction of the corner brown building, where both agents independently construct the identical geometric shape of the architecture from completely different approach angles.}
\label{fig:cross_agent}
\end{figure*}

We further examine the generation results across multiple interactive topologies in Fig. \ref{fig:cross_agent}. This analysis verifies environmental alignment through two complementary dimensions: 2D video frame consistency and 3D point cloud stability.

Multi-agent 2D frame consistency. ShareVerse effectively maintains cross-agent visual consistency across different vehicle perspectives. The generated 4-view video frames exhibit high cross-agent visual consistency for interacting vehicles. By comparing the respective grids, it is clear that specific scene elements, such as storefront architectures, remain structurally identical across different vehicle perspectives.

Geometric 3D point cloud consistency. To verify that the generated world is geographically anchored, we perform independent 3D point cloud reconstructions for each agent. As shown at the bottom of each panel in Fig. \ref{fig:cross_agent}, the 3D topologies reconstructed from independent sequences are nearly identical across all evaluated scenarios. This accurate alignment indicates that the model synthesizes a stable physical space rather than merely visually similar frames. Even in the highly challenging X-junction crossing (Fig. \ref{fig:cross_agent}f), the consistent 3D reconstruction of the intersection from different approach angles demonstrates an accurate understanding of the shared coordinate system, as further detailed in the figure caption.

\begin{figure*}[h] 
    \centering
    \includegraphics[width=\linewidth]{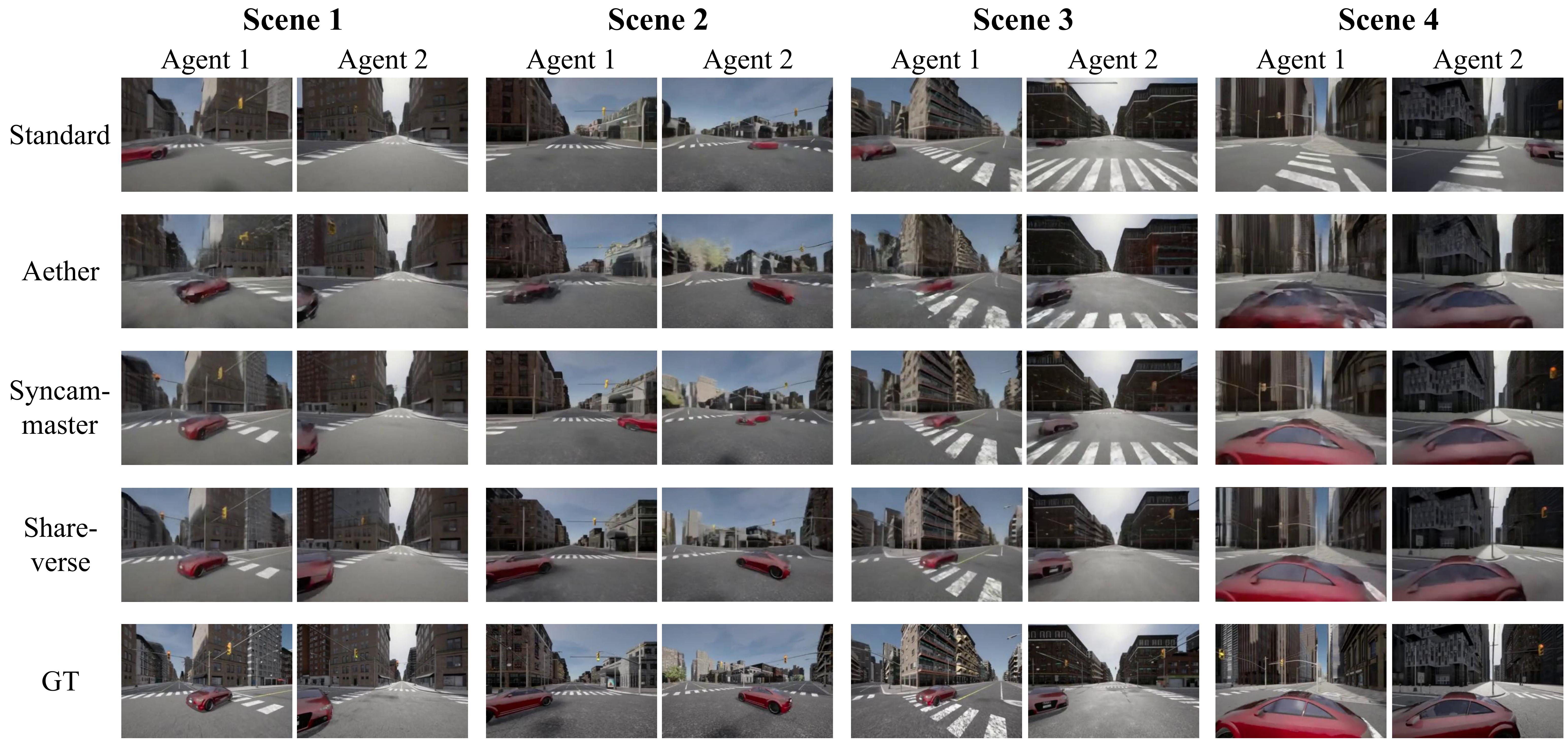}
    \vspace{-6mm}
    \caption{\textbf{Qualitative comparison of multi-agent dynamic object consistency during mutual visibility.} 
Each row displays the generated video frames from the specific camera views where the two agents observe each other at the intersection moment.
The Ground Truth results are displayed as a reference (row 5). 
The comparison focuses on whether interacting agents can faithfully perceive each other. 
Crucially, ShareVerse (row 4) accurately renders the specific red vehicle (the other agent) with high visual fidelity in the correct spatial location, whereas other baselines fail to synthesize the dynamic object consistently. 
Note that while the generated frames differ slightly from the Ground Truth reference due to inherent stochasticity in video generation, ShareVerse still maintains superior frame quality and cross-agent consistency.}
  \vspace{0mm}
\label{fig:cross_agent_dynamic}
\end{figure*}

Dynamic cross-agent consistency. A truly shared world requires strict synchronization not only for static architectures but also for dynamic entities. Figure \ref{fig:cross_agent_dynamic} compares ShareVerse against baselines during a critical interactive event: two agents passing each other. A valid shared physical space demands mutual visibility, meaning Agent 1 and Agent 2 must simultaneously observe one another in their respective viewports. While baseline methods frequently fail to synthesize the counterpart vehicle or produce spatial misalignments, ShareVerse successfully maintains this dynamic mutual observation. By accurately rendering the interacting vehicles at the shared spatiotemporal coordinates, ShareVerse extends its consistent spatial logic to dynamic scenarios.



\subsection{Quantitative Evaluation of Video Generation Fidelity.} 
Following the qualitative analysis, Fig. \ref{fig:comparison}(f) visualizes the quantitative reconstruction metrics. 
ShareVerse consistently outperforms all baseline methods, achieving a PSNR of 21.76, an SSIM of 0.6637 and an LPIPS of 0.2570. 
This appreciable drop in pixel error and perceptual error quantitatively corroborates the performance of the reprojection-based metric mentioned in Sec. \ref{cross-agent consistency} and the structural alignment observed in the visual comparisons. 
Ultimately, the numerical improvements demonstrate ShareVerse's strong capability to minimize structural discrepancies and maintain an accurate shared environment across distinct agent views.

\section{Ablation Study}
To further verify the effectiveness of our method, we conduct ablation studies, as shown in Table \ref{tab:ablation}. We compare training on four-view videos with that on single-view(front) videos, with metrics computed only on front-view videos at the same resolution. The results show that the former performs better. Due to the significant divergence in the observational perspectives of the two interacting agents, a single viewpoint alone is insufficient to acquire full-scene information, which greatly hinders the transmission and sharing of visual information between them.

With respect to model design, the cross-agent attention module unsurprisingly plays an important role in interactive generation.

\begin{figure}[h]
	\centering
\includegraphics[width=1\linewidth]{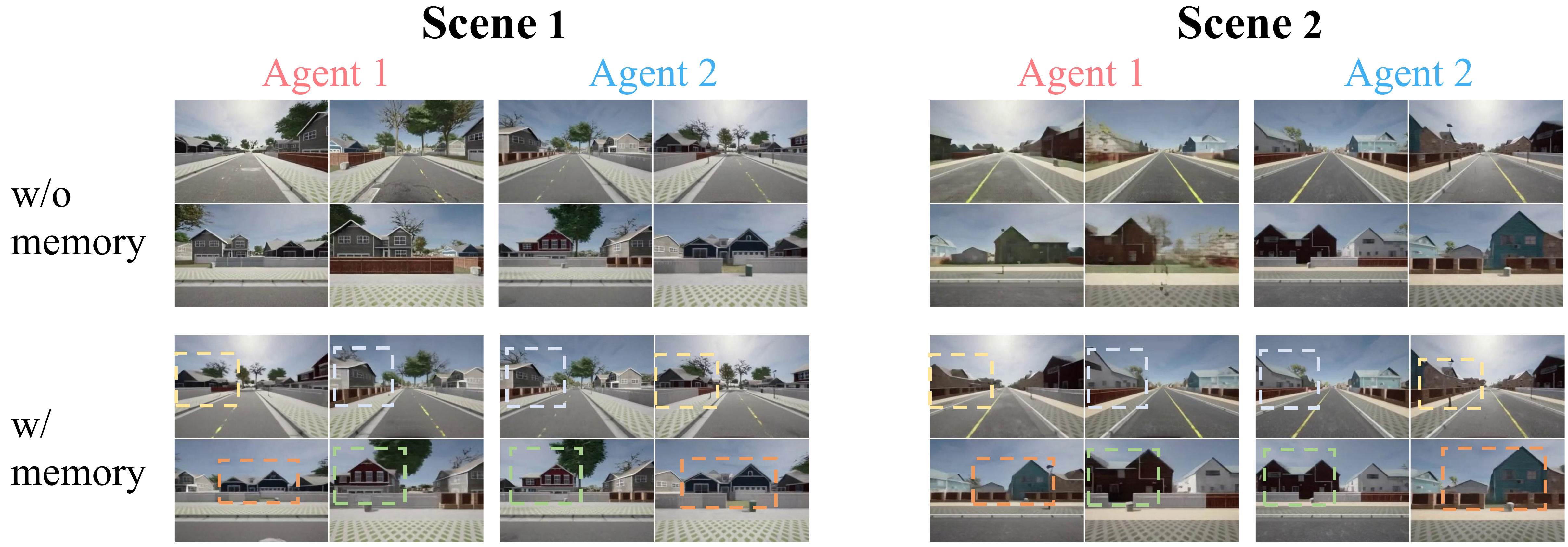}
    \vspace{-6mm}
    \caption{The visualization of the memory mechanism’s effect. }
 \label{Fig:ablation}
\end{figure}

In addition, experiments are conducted on autoregressive generation with and without memory. Undoubtedly, the memory mechanism ensures long-range consistency, critical to shared world modeling.Some visual examples are shown in Fig. \ref{Fig:ablation}.

\begin{table}[htbp]
 \centering
 \begin{tabular}{lccc}
 \toprule
 \textbf{Method} & \textbf{PSNR} $\uparrow$ & \textbf{SSIM} $\uparrow$ & \textbf{LPIPS} $\downarrow$ \\
 \midrule
 \multicolumn{4}{c}{\textbf{Full sequence}} \\
 \midrule
 Front view only          &20.51  &0.6762  &0.3230  \\
 Four views concat         &\textbf{22.28}  &\textbf{0.6862}  &\textbf{0.2258}  \\
  \cmidrule{2-4}
 w/o cross-agent attn     &20.42  &0.6304  &0.2935  \\
 Full sequence (Ours)     &\textbf{21.76}  &\textbf{0.6637}  &\textbf{0.2570}  \\
 \midrule
 \multicolumn{4}{c}{\textbf{Autoregressive}} \\
 \midrule
 w/o memory               &19.58  &0.6132  &0.3443  \\
 Autoregressive (Ours)    &\textbf{20.71}  &\textbf{0.6439}  &\textbf{0.2929}  \\
 \bottomrule
 \end{tabular}
 \caption{Ablation Study on reference-based metrics.}
 \label{tab:ablation}
\end{table}

\section{Conclusion}
ShareVerse defines a novel paradigm for multi-agent shared world modeling, ensuring accurate geometric and visual consistency across independent entities by spatially concatenating multi-view videos, integrating cross-agent attention blocks and employing a memory retrieval mechanism. By bridging the gap between isolated observations and a synchronized environment, it establishes a robust foundation for collaborative embodied AI. Future work will target real-time, highly dynamic physical interactions.


\bibliographystyle{ACM-Reference-Format}
\bibliography{sample-base}

\appendix









\end{document}